\documentclass[conference]{IEEEtran}

\IEEEoverridecommandlockouts
\usepackage{cite}
\usepackage{amsmath,amssymb,amsfonts}
\usepackage{algorithmic}
\usepackage{graphicx}
\usepackage{textcomp}
\def\BibTeX{{\rm B\kern-.05em{\sc i\kern-.025em b}\kern-.08em
    T\kern-.1667em\lower.7ex\hbox{E}\kern-.125emX}}
\begin{document}

\title{A Model for Medical Diagnosis \\
Based on Plantar Pressure\\
\thanks{This paper is partially supported by the National Natural Science Foundation of China (NSFC Grant Nos. 91646202, 61772039 and 61472006) and Peking University Seed Fund for Medicine-Information Interdisciplinary Research Project.
}
}

\author{\IEEEauthorblockN{1\textsuperscript{st} Guoxiong Xu}
\IEEEauthorblockA{\textit{School of Electronics Engineering and Computer Science} \\
\textit{Peking University}\\
Beijing, China \\
xgx@pku.edu.cn} \\
\IEEEauthorblockN{3\textsuperscript{rd} Hongshi Huang}
\IEEEauthorblockA{\textit{Institute of Sports Medicine} \\
\textit{Peking University Third Hospital}\\
Beijing, China \\
13910093298@163.com} \\
\IEEEauthorblockN{5\textsuperscript{th} Can Liu}
\IEEEauthorblockA{\textit{School of Electronics Engineering and Computer Science} \\
\textit{Peking University}\\
Beijing, China \\
vvoliucano@gmail.com} \\
\and
\IEEEauthorblockN{2\textsuperscript{nd} Zhengfei Wang}
\IEEEauthorblockA{\textit{School of Electronics Engineering and Computer Science} \\
\textit{Peking University}\\
Beijing, China \\
wangzhengfei@pku.edu.cn} \\
\IEEEauthorblockN{4\textsuperscript{th} Wenxin Li}
\IEEEauthorblockA{\textit{School of Electronics Engineering and Computer Science} \\
\textit{Peking University}\\
Beijing, China \\
lwx@pku.edu.cn} \\
\IEEEauthorblockN{6\textsuperscript{th} Shilei Liu}
\IEEEauthorblockA{\textit{School of Electronics Engineering and Computer Science} \\
\textit{Peking University}\\
Beijing, China \\
pkucs.lsl@gmail.com} \\
}

\maketitle

\begin{abstract}
The process of determining which disease or condition explains a person's symptoms and signs can be very complicated and may be inaccurate in some cases. The general belief is that diagnosing diseases relies on doctors' keen intuition, rich experience and professional equipment. In this work, we employ ideas from recent advances in plantar pressure research and from the powerful capacity of the convolutional neural network for learning representations. Here, we propose a model using convolutional neural network based on plantar pressure for medical diagnosis. Our model learns a network that maps plantar pressure data to its corresponding medical diagnostic label. We then apply our model to make the medical diagnosis on datasets we collected from cooperative hospital and achieve an accuracy of 98.36\%. We demonstrate that the model base on the convolutional neural network is competitive in medical diagnosis.
\end{abstract}


\section{Introduction}
Diagnosis is often challenging as many symptoms and signs are nonspecific. For example, redness of the skin is a sign of many disorders and thus cannot tell diagnosis. Therefore doctors must perform some differential diagnosis to compare and contrast several possible explanations. This process requires experienced medical experts and precise equipment, which leads to three adverse effects:
\begin{itemize}
\item Over-diagnosis: Diagnosing an innocent person without disease as a patient. This can result in economic waste and unnecessary treatments may cause harm.
\item Errors: Also referred to as medical errors, errors might include an inaccurate or incomplete diagnosis or treatment of a disease, injury or ailment. This is evidently harmful to the patient indeed.
\item Lag time: Lag time is the interval between start and end of the diagnosis of a disease or condition. Overlong lag time may lead to exacerbation of the circumstance.
\end{itemize}

All adverse effects can be improved obviously by more experienced medical experts, more advanced equipment and more efficient approach. However, the number of experts is always limited and it is impossible to give everyone the chance to be diagnosed by experts. Besides, advanced equipment always means complexity and high cost, which leads to many chain problems. Therefore, in this work, our focus is on the last one, i.e. approach to mapping observations to concrete diagnosis.

Plantar pressure refers to pressure fields acting between the plantar surface of the foot and a supporting surface. It is employed in a wide range of applications including sports biomechanics and gait biometrics. One of the earliest recorded attempts to measure was made by Beely in 1882 \cite{b1}. The hardware to measure plantar pressure currently fall into two main categories: floor-based and in-shoe. With the development of plantar pressure measurement, the research and application thrive meanwhile.

In clinical research and application, much work concentrate on the foot and lower limb disease's impact on plantar pressure. Prior studies focus on diseases like diabetic foot \cite{b2}, anterior cruciate ligament injury \cite{b3} and Parkinson disease \cite{b4}. These research show that patients with diseases above may change their walking pattern to some degree to avoid sore and pain, which leads to the distribution alteration of plantar pressure. Previous typical approaches \cite{b5}\cite{b6}\cite{b7}\cite{b8}\cite{b9} to analyze plantar pressure try to find the characteristics as much as possible. All these characteristics present the original plantar pressure circumstance in many ways and used to further diagnosis.

The novelty of our work is that we propose a model for medical diagnosis using plantar pressure, which fully utilizes convolutional neural networks' powerful ability to learn representations of plantar pressure. We demonstrate that our model is capable of medical diagnosis on datasets we collected from our cooperative hospital, making it more accurate and convenient to make a diagnosis.

Besides our contribution in improving the accuracy and usability of medical diagnosis, we contribute by proposing and verifying a brand-new solution to medical data. We hope that our work will improve and benefit medical domain.

We organize the paper by first briefly elaborating on some of the related work to the task and our model. Then in the following section, we introduce and explain our model. In Section IV we describe our work on data collection and experiments we performed, demonstrating strong results on medical diagnosis using our model based on collected datasets.

\section{Related Work}
Though it may be impossible to say that people can make the medical diagnosis only based on data from feet, many research \cite{b2}\cite{b3}\cite{b4}\cite{b17}\cite{b18} have demonstrated the feasibility of this approach. According to these work, people find out some relationships between plantar pressure data and diseases like diabetic foot \cite{b2}, anterior cruciate ligament injury \cite{b3} and Parkinson disease \cite{b4}. With some analysis of plantar pressure data, researchers can make a concrete medical diagnosis even without any medical knowledge.

As is discussed in Section I, previous approaches to analyze plantar pressure were trying to find the characteristics to represent the original data. A characteristic stands for a corresponding feature in one aspect. With characteristics as an auxiliary, medical experts can make diagnosis more accurate.

Different plantar pressure data varies in direction and foot size. A widely recognized normalization method \cite{b6} is to normalize with a characteristic called foot progression angle (FPA). This method divides the original plantar pressure data with a critical value and extract common tangent from both forefoot and rearfoot, then defines FPA as the average angle of the tangent lines to the medial and lateral side of the foot \cite{b6}. This ensures subsequent process to be performed on the same standard.

There are many characteristics proved to be effective: arch index (AI) \cite{b9}, which is usually used to indicate flat foot and high arch foot. Heel, medial midfoot (MM), medial forefoot (MF), lateral midfoot (LM), lateral forefoot (LF) are a standard of regional division, and percent medial impulse (PMI) which is calculated through above regional division reflects the impulse percentage exerted on the medial aspects \cite{b7}. Trajectory of the center of pressure (COP) \cite{b5} represents the movement of body's center of gravity. And peak pressure (PP), mean pressure (MP), pressure-time integral (PTI) \cite{b8} are three ways to represent a certain characteristic. These characteristics describe a plantar pressure data in different ways according to its original distribution.

The work in medical diagnosis using plantar pressure take advantage of characteristics extracted under methods described above manually. They combined separate information together using complex equations to find hidden relationships between them and to make accurate and concrete diagnosis. For example, work in \cite{b17} and \cite{b18} calculate various characteristics following those methods from original plantar pressure data and then pass them to fully connected neural network based classifier. These research find that disease can result in abnormal distribution of plantar pressure, and those characteristics are significantly different.

\section{Model}
Our model to make a medical diagnosis using plantar pressure is based on the work which we describe in the previous section. Those work still rely on manually extracted characteristics. From our point of view, there is more information hidden in plantar pressure data and people cannot describe or even find out them at all. In other words, we try to find characteristics as many as possible to represent the original plantar pressure data.

We draw inspiration from the recent breakthrough of deep learning \cite{b10} and powerful ability of convolutional neural network (CNN) to learn representations from data effectively and automatically \cite{b11}. Compared with prior work, we employ CNN to learn representations from plantar pressure, which we argue that CNN can find out information people impossibly notice.

Our model is divided into portions we describe in the following subsections: data preprocessing, representations learning and voting mechanism. First, we deal with the raw plantar pressure data with picking the maximum, calculating the sum and average to present the data in different ways. Second, we train a CNN model to learn the representations for each type. Finally, we make the medical diagnosis according to every model's voting result.

\subsection{Data Preprocessing}
Plantar pressure data consist of many frames, every frame contains some data points and reflects the instantaneous state that how the foot contact with the ground. Many works concentrate on how to process these data points. Prior works \cite{b5}\cite{b6}\cite{b7}\cite{b8}\cite{b9} create many computing methods to deal with the original plantar pressure data and these methods describe original data in many aspects. Inspired by work in \cite{b8}, we propose a new method to preprocess the original data. Based on this method, we can take full advantage of CNN's representation learning ability. Given raw plantar pressure data, we present it with maximum data points, sum and average of all data points over frames.

This method is inspired by definition of peak pressure (PP), mean pressure (MP), pressure-time integral (PTI) in \cite{b8}. PP describes the maximum of every location's pressure, while MP describes the average circumstance. And PTI shows the integral of pressure at every data point. In a word, we process the raw data to save their maximum, sum and average form.

By picking the max data point, calculating the sum and average, we can present the original data in different ways, which we argue that will reduce the loss in preprocessing. Each type describes the data in a specific way, making up for other types' information loss.

Given a raw plantar pressure data ${D_{raw}}$, which consists of ${K}$ frames data. ${{d_{raw}}^k_{i,j}}$ stands for the data point with the coordinate ${(i,j)}$ in the ${k}$-th frame data, and ${{K'}_{i,j}}$ counts the total number of frames where the corresponding position data is not zero. Based on above definition, ${D_{max}}$ for data of maximum, ${D_{sum}}$ for data of sum and ${D_{average}}$ for data of average can be written as:

\begin{equation}
d_{max_{i,j}}=max({d_{raw}}^k_{i,j})\label{eq}
\end{equation}

\begin{equation}
d_{sum_{i,j}}=\sum^K_{k=1}{d_{raw}}^k_{i,j}\label{eq}
\end{equation}

\begin{equation}
d_{average_{i,j}}=\frac{1}{{K'}_{i,j}}\sum^K_{k=1}{d_{raw}}^k_{i,j}\label{eq}
\end{equation}

Then we convert these data matrices into grayscale images, the pixel will get closer to white as the corresponding data point's value is larger. We also normalize the images into uniform sizes for subsequent use according to the measuring equipment specification. The process is shown in Figure 1. 

This data preprocessing method can deal with any plantar pressure data automatically, and transform the raw data into images which can be dealt with automatically by standard CNNs. Then we start processing these images data and looking for hidden information in them.

\begin{figure}[htbp]
\centerline{\includegraphics[scale=.75]{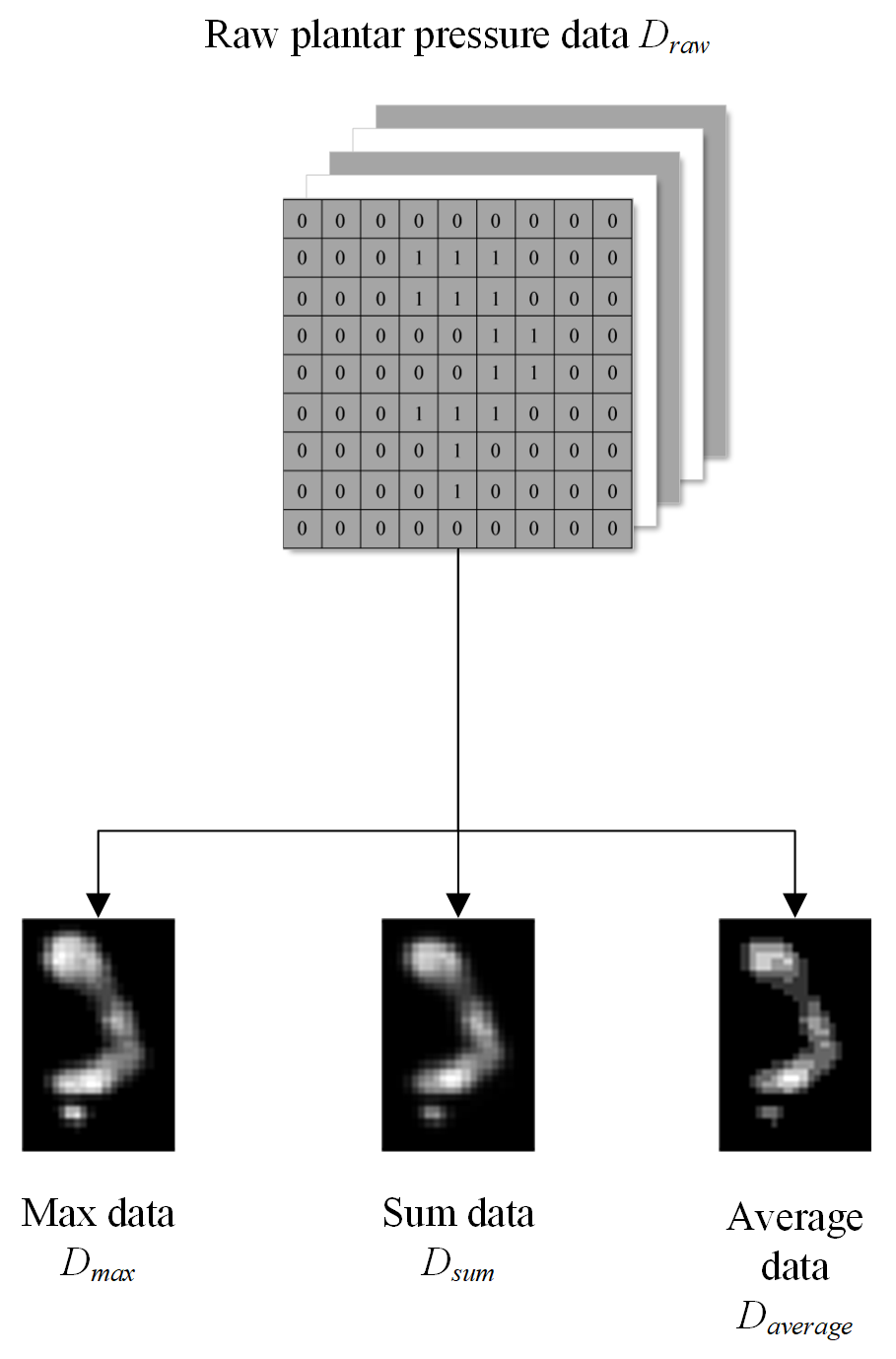}}
\caption{Data preprocessing. Pick the maximum data point, calculate the sum and average of data points over frames. Then convert them to grayscale images.}
\label{fig1}
\end{figure}

\subsection{Representation Learning}
Convolutional neural networks have achieved exceptional results in many large-scale computer vision applications, particularly in image recognition task \cite{b12}\cite{b13}\cite{b14}\cite{b15}. In 2015, ResNet \cite{b15} exceeded human-level accuracy in image recognition. Since then, more focus is now being placed on more challenging components, such as object detection and localization. In 2016, research \cite{b16} shows that CNN has the ability to control an autonomous vehicle and the ability to learn representations from images data automatically. Considering this powerful ability to learn representations automatically, we employ that using CNN to find the characteristics from the preprocessed images data.

According to work in \cite{b11}, with several convolutional layers linearly connected, a neural network can find out useful geometrical characteristics to represent the original images. We use convolutions to find characteristics in plantar pressure images data automatically and then feed these representations into fully connected layers to classify.

We propose a CNN model whose architecture is shown in Figure 2. The model consists of 7 layers, including a normalization layer, 4 convolutional layers and 2 fully connected layers. The convolutional layers are designed to perform representations extraction and are chosen empirically through a series of experiments with varies layer configurations, and the fully connected layers can be seen as diseases classifier. The input image is one of three representing method of raw plantar pressure data. We use different type of data to train corresponding model and therefore get three models with the same architecture and different parameters.

\begin{figure}[htbp]
\centerline{\includegraphics[scale=.7]{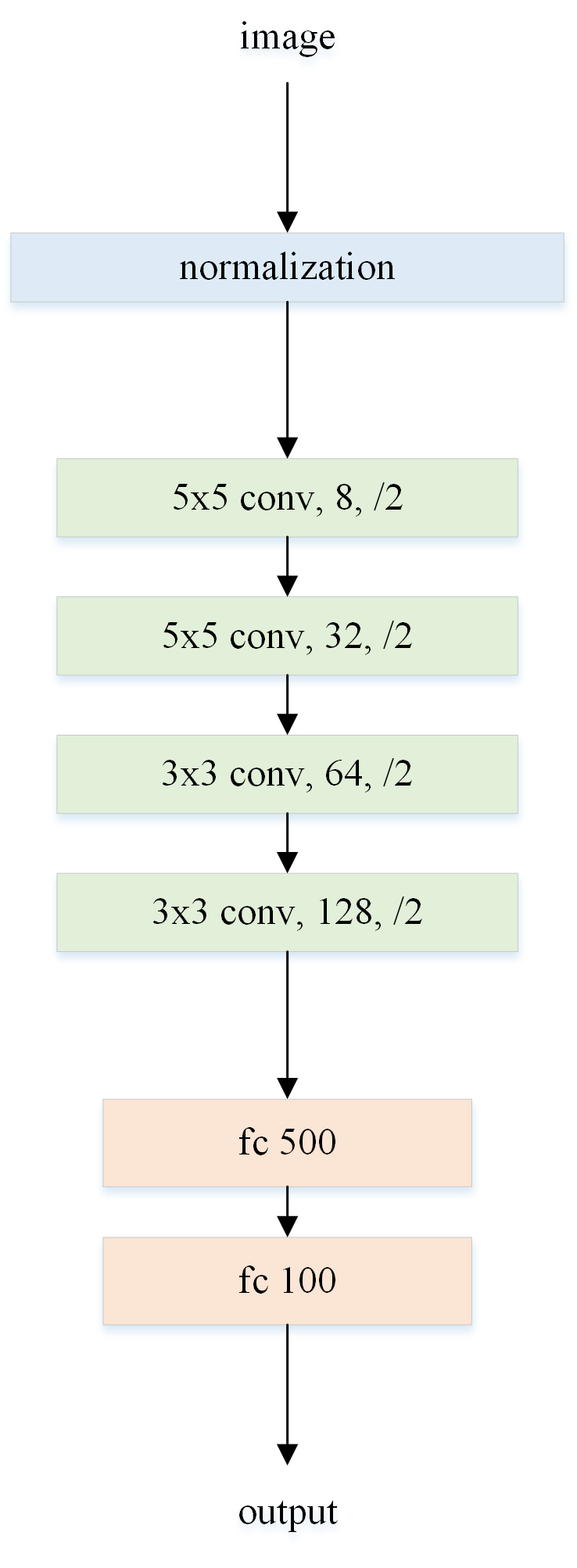}}
\caption{Convolutional neural network architecture. The network has about 900 thousand parameters.}
\label{fig2}
\end{figure}

\subsection{Voting Mechanism}
With trained CNN models, we classify each type of data successfully and make the medical diagnosis using plantar pressure data more accurately. Every CNN model analyzes the data and classifies under some regulations it learns automatically for a corresponding type. However, as we deal with the raw data with described preprocessing method, there must be some loss.

\begin{figure}[htbp]
\centerline{\includegraphics[scale=.75]{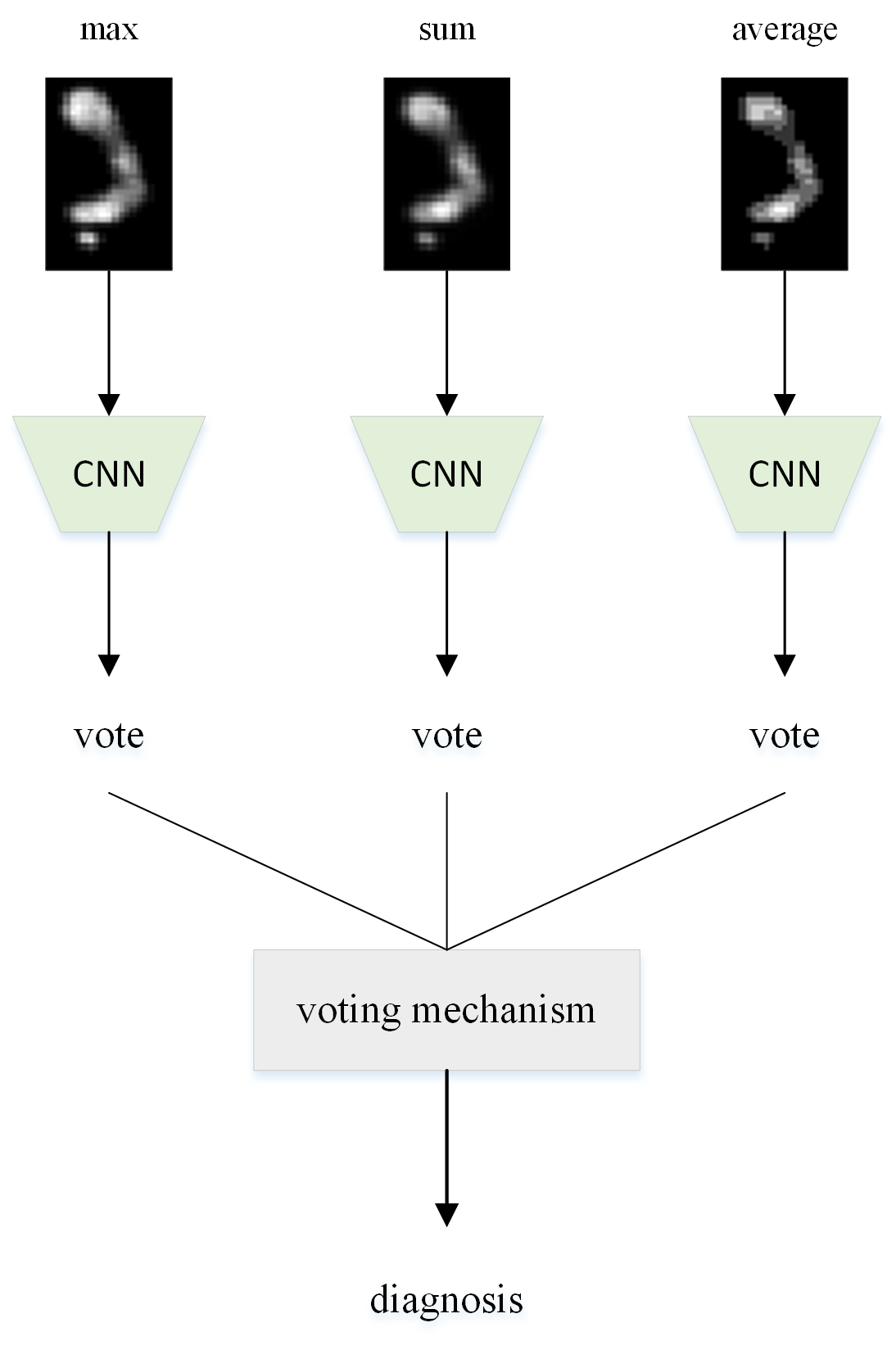}}
\caption{Voting mechanism. Each CNN votes according to its computation towards the corresponding image. The final diagnosis can be obtained following the principle of majority rule.}
\label{fig3}
\end{figure}

In order to guarantee the accuracy and reliability of our model, we should adopt some mechanism to deal with the loss in preprocessing procedure. Therefore, we combine our CNN models together for a single case to reduce information loss and improve accuracy, which we call voting mechanism.

Figure 3 shows a block diagram of our voting mechanism. For a single case, we feed different type of data into corresponding CNN model to compute a medical diagnosis. Then we let these CNN models "vote" for the final diagnosis and we can get the result following the principle of majority rule.

\section{Data Collection \& Experiments}
In this section, we describe our work on data collection in cooperation with hospital and experiments verifying our model performance.

\subsection{Data Collection}
A measuring equipment is required for the data collection work. The equipment applied is a plate with sensor matrix to measure the entire plantar pressure during the measurement. Volunteers need to walk casually on the measurement. To avoid the influence of shoes and socks, our collection required volunteers keep foot bare during the measurement. Additionally, volunteers should walk around at least five times to ensure the accuracy and stability of collected data.

We arrange this equipment in our cooperative hospital to collect data. Our targeted disease this time is anterior cruciate ligament (ACL) deficiency, a debilitating sports injury that leads to altered knee loading, affecting performance of daily living activities and increasing the risk of early osteoarthritis of the knee \cite{b19}\cite{b20}, which has been proved to affect plantar pressure \cite{b3}.

So far, we have 129 volunteers in total, 64 of them are patients suffering ACL deficiency and remaining are volunteers without this disease. Each volunteer's data consists of 5 bare walking plantar pressure data for both left and right foot. With the high sampling frequency and high density of sensors, the equipment can record over 150 frames for each data and every frame contains about 1,000 data points.

It is worth to mention that this equipment is still placed at our cooperative hospital, what we expect is to collect more plantar pressure data and information of patients with various diseases, not only ACL deficiency, for further research.

\subsection{Experiments}
All our experiments revolve around the same basic task: diagnosing ACL deficiency from our collected plantar pressure data.

We consider one-time successful measurement as an independent case, after preprocessing the raw datasets, we get 645 cases in total, which contains 320 cases with ACL deficiency and 325 cases without. Each case consists of 6 data (images): max left, max right, sum left, sum right, average left and average right. We divide the whole cases into two parts, 145 cases from 14 patients and 15 healthy persons are used as the test sets, and remaining 500 cases from 50 patients and 50 healthy persons are used as training sets for the CNN models.

For ACL deficiency diagnosis, we considered a baseline that calculates characteristics from plantar pressure data using traditional methods and then feeds these characteristics into a machine learning model trained by support vector machine (SVM) algorithm, which is the typical way to classify plantar pressure data in the past.

\begin{table}[htbp]
\caption{Result on collected ACL deficiency datasets}
\begin{center}
\begin{tabular}{|c|c|c|c|c|c|}
\hline
\textbf{Model} & \textbf{Baseline} & \multicolumn{4}{|c|}{\textbf{Our model categories$^{\mathrm{a}}$}} \\
\cline{2-6}
\textbf{Type} & \textbf{SVM} & \textbf{\textit{max}}& \textbf{\textit{sum}}& \textbf{\textit{average}}& \textbf{\textit{voting}} \\
\hline Accuracy& 76.67\%& 94.67\%& 92.62\%& 88.93\%& \textbf{98.36}\% \\
\hline
\multicolumn{6}{l}{$^{\mathrm{a}}$}Max, sum, average and vote refer to methods described in Section III.
\end{tabular}
\label{tab1}
\end{center}
\end{table}

Result comparing the baseline to our model are shown in Table I. Each type of CNN model can reach an accuracy around 90\%, outclassing baseline of finding characteristics manually. Furthermore, the model using our proposed voting mechanism improves the accuracy from about 90\% to amazing 98.36\%. This outstanding performance demonstrates that the model we proposed using CNN and voting mechanism is an effective way to diagnose disease related to plantar pressure like ACL deficiency.

\subsection{Discussion}
It is reasonable that our model outperforms than the SVM method. As is previously described, traditional methods rely on extracting characteristics manually. Generally speaking, given more features, SVM can get better performance, therefore, the only path to improve the accuracy of SVM is to find out more and more features from original plantar pressure data. However, it is impossible for human to analyze the whole information from an image using eyes. On the contrast, our model could analyze the images and find relationships between extracted information and ground-truth label automatically, which in our opinion makes our model outperforms.

Another interesting point shown from the result is the sum-model performs much better than average-model. According to equation (3), the average data is not simply dividing sum data with a constant. To put it in mathematical way, the sum data can be regarded as the integral of plantar pressure over whole walking time, while the average data only reflects the mean state when the sensors are being pressed. From the result, the sum-model's higher accuracy shown that there are different information hidden in the whole data and they reflect the plantar pressure from different aspects. The voting mechanism's highest performance proves that they can make up for each other.

\section{Conclusion}
In this paper, we introduce a new convolutional neural network based model which is capable of making medical diagnosis for lower limb disease. This work demonstrates that convolutional neural network can also learn the representations in the medical images or data field automatically with great effect. Further, we propose a preprocessing method for plantar pressure to convert mysterious original data to more understandable data. An obvious drawback of our work is the fact that, the collected data only differ in volunteers with and without anterior cruciate ligament deficiency. Under our ideal assumptions, all lower limb diseases can be analyzed from plantar pressure data. Much of our future efforts will concentrate on collecting more data with more kinds of lower limb diseased and more analysis on them. Besides, we will try to make our whole plantar pressure data open source to attract more people involved in this area. We feel this is an area with exciting challenges which we hope to keep improving in future work.

\end{document}